%% file: ms.tex
\documentclass[a4paper, 10pt, conference]{ieeeconf}      % Use this line for a4 paper
\usepackage{newclude}
\IEEEoverridecommandlockouts                              % This command is only needed if 
\input{preambel}
\usepackage[noadjust]{cite}

\makeatletter
\let\NAT@parse\undefined
\makeatother

\usepackage[colorlinks=true,
 			linkcolor=red!85!blue,
 			citecolor=black!30!green,
 			urlcolor=red!25!blue,
 			breaklinks=true,
            pagebackref=true,
]{hyperref}
\usepackage[capitalise]{cleveref}

\input{title_and_author}

\begin{document}

\maketitle
\thispagestyle{plain}
\pagestyle{plain}

\include*{abstract}
\include*{introduction}
\include*{related_work}
\include*{method}
\include*{experiments}

\include*{conclusion_and_acknowledgments}

\bibliographystyle{IEEEtran}
\bibliography{IEEEabrv,bibliography}

\end{document}

%% file: preambel.tex
\usepackage{siunitx}
\usepackage{amsmath}
\usepackage[utf8]{inputenc} % allow utf-8 input
\usepackage{url}            % simple URL typesetting
\usepackage{booktabs}       % professional-quality tables
\usepackage{amsfonts}       % blackboard math symbols
\usepackage{nicefrac}       % compact symbols for 1/2, etc.
\usepackage{microtype}      % microtypography
\usepackage{graphicx}
\usepackage{subcaption}
\usepackage{pgfplots}
\usetikzlibrary{pgfplots.groupplots}
\pgfplotsset{compat=1.14}

\newcommand*{\norm}[1]{\left\lVert#1\right\rVert}		% Norm
\newcommand*{\w}{\omega}
\newcommand*{\E}{\mathbf{E}}
\newcommand*{\HH}{\mathbf{H}}
\newcommand*{\I}{\mathbf{I}}
\newcommand*{\D}{D_{\text{train}}}
\newcommand*{\KL}{\text{KL}}
\newcommand*{\Var}{\text{Var}}

%% file: title_and_author.tex
\title{\LARGE \bf
Uncertainty Estimation in One-Stage Object Detection
}

\author{Florian Kraus$^{1}$ and Klaus Dietmayer$^{2}$% <-this % stops a space
\thanks{$^{1}$Florian Kraus is with Daimler AG, 89081 Ulm, Germany,
}%
\thanks{$^{2}$Klaus Dietmayer is with Institute of Measurement, Control and Microtechnology, Ulm University, 89081 Ulm, Germany}%
}

%% file: abstract.tex
\begin{abstract}
Environment perception is the task for intelligent vehicles on which all subsequent steps rely. A key part of perception is to safely detect other road users such as vehicles, pedestrians, and cyclists. With modern deep learning techniques huge progress was made over the last years in this field. However such deep learning based object detection models cannot predict how certain they are in their predictions, potentially hampering the performance of later steps such as tracking or sensor fusion. We present a viable approaches to estimate uncertainty in an one-stage object detector, while improving the detection performance of the baseline approach. The proposed model is evaluated on a large scale automotive pedestrian dataset. Experimental results show that the uncertainty outputted by our system is coupled with detection accuracy and the occlusion level of pedestrians.
\end{abstract}

%% file: introduction.tex
\section{Introduction}
\label{sec:introduction}
Object detection is a crucial task for safe autonomous driving. With the introduction of deep learning the performance of object detection made a huge leap forward. However most of these methods have no measure of how certain they are in their output. When confronted with previously unseen data there is usually no way to measure if the model can deal with this input. For example a model trained on good weather data is faced with adverse weather situations.

Capturing uncertainty in object detection can provide a measure of how certain a model is in its output. Providing an uncertainty measure could also improve subsequent steps such as sensor fusion and tracking or be beneficial for active learning. Annotating images for object detection is a time consuming and costly process. To label only the data with the most information gain could help cutting time and costs.

Bayesian methods have a long history of providing an uncertainty measure. Bayesian Neural Networks (BNNs) are one way to apply Bayesian concepts into neural networks. In a BNN each weight is a random variables. As a result the output of such a model is also a random variable, providing a way to measure uncertainty.

There are two distinct concepts of uncertainty, \emph{epistemic} and \emph{aleatoric}. Epistemic uncertainty, is uncertainty which can be explained away given more data. This measures model uncertainty and is captured by Bayesian methods which become more confident the more data is used to train them.
Aleatoric uncertainty, on the other hand, is data or problem inherent and cannot be reduced with more data. This includes sensor noise or ambiguities in the problem itself, e.g. matching problems in stereo depth map generation. It can be captured by explicitly modeling it as a model output.

Until recently it was not practical to apply Bayesian methods to neural networks. Gal and Ghahramani \cite{Gal2016Bayesian} showed that a BNN can be approximated by incorporating dropout into an ordinary neural network. Kendall and Gal \cite{kendall_what} also introduced aleatoric uncertainty measures predicted by the model itself. They applied both epistemic and aleatoric uncertainty estimation to instance and scene segmentation and depth estimation.

\begin{figure}[t]
    \centering
    \includegraphics[width=\columnwidth]{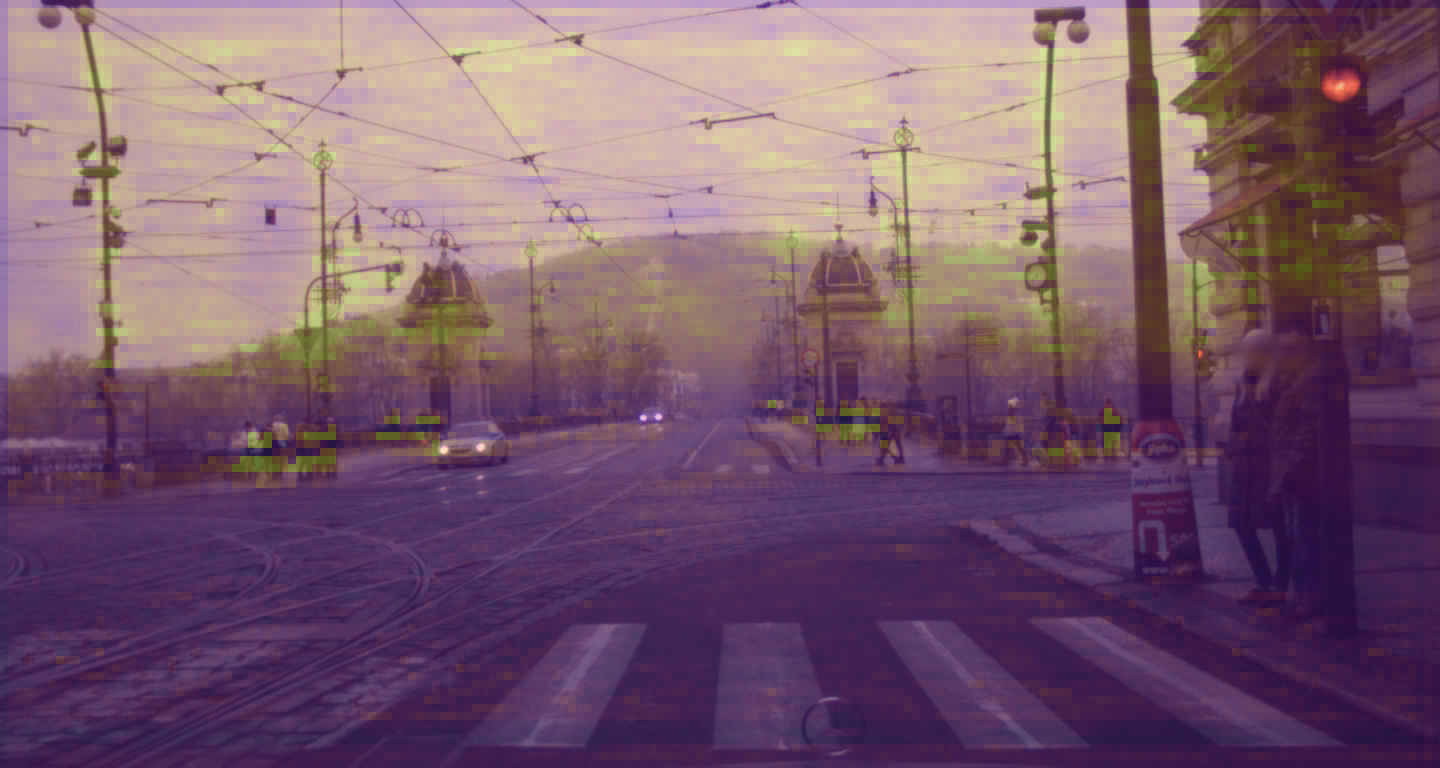}
    \caption{Aleatoric variance for bounding box offset in $y$ (vertical) direction. Variance is shown for prior box size of 55 by 20 pixels (height, width). Bright yellow indicates high uncertainty relative to the background. Notice the high variance for predictions which are slightly off vertically.}
    \label{fig:title}
\end{figure}
Recently their methods were incorporated into object detection by \cite{di_feng_first}, \cite{wirges2019capturing}. Both used two-stage models which first generate proposals from a region proposal network and, in a second network, refine the box offsets and predict a class. Incorporating dropout into an one-stage object detector to capture uncertainty was first done by \cite{miller2018sampling}. However, they did not use the methods provided by \cite{Gal2016Bayesian} to approximate the predictive distribution of the BNN, making their approach similar to model ensembles, which are an orthogonal method of capturing uncertainty. Our approach on the other hand uses the methods of \cite{Gal2016Bayesian} to approximate the predictive distribution.

In this article we rigorously incorporate the methods provided by \cite{Gal2016Bayesian} and \cite{kendall_what} into an one-stage object detection framework. We provide all the necessary theory to do so and make our code publicly available\footnote{\url{https://github.com/flkraus/bayesian-yolov3}}. Our model is a reimplementation of YOLOv3 \cite{redmonYOLOv3} in TensorFlow \cite{tensorflow2015-whitepaper} with an adjusted loss function to accommodate the Bayesian setting. The here presented loss function and techniques can also be applied to other network architectures and are not tied to the YOLOv3 architecture.
We apply our method to train a pedestrian detector on an automotive pedestrian dataset and carry out experiments to compare the different uncertainty estimations. Furthermore we discuss the effects of using an one-stage object detector instead of a two-stage approach.

%% file: related_work.tex
\section{Related Work}
\label{sec:related_work}

Since \cite{Gal2016Bayesian} provided a straightforward way to incorporate Bayesian concepts into neural networks, there have been numerous applications of these ideas. In the context of this article we are interested in applications to object detection.

Both one-stage and two-stage object detection methods have their advantages. Two-stage detectors typically have better performance where one-stage detectors have better run time making them suitable for automotive applications \cite{Huang_2017_CVPR}.

Uncertainty estimation for two-stage object detection was previously evaluated by Feng et al. \cite{di_feng_first} and Wirges et al. \cite{wirges2019capturing}. Both applied the methods proposed in \cite{kendall_what} to a two-stage network for object detection in 3D Lidar data. They tested their models on the KITTI dataset \cite{Geiger2013IJRR}.
Dropout was applied to the fully connected layers after the region proposal network (RPN). Additionally, \cite{wirges2019capturing} added dropout to some of the convolutional layers of the RPN. When applying dropout to all layers of the RPN, they found that the model no longer converges.
\cite{di_feng_first} found that detections which had a high Intersection over Union (IoU) with a ground truth object had lower epistemic uncertainty than those with lower IoU scores. For aleatoric uncertainty they found that the aleatoric uncertainty was correlated with the distance of objects.
\cite{wirges2019capturing} included the pedestrian and cyclist classes into their training alongside the car class.
They found that cars have a lower epistemic uncertainty, which, they argued, could be due to the under representation of the pedestrian and cyclist classes
compared to the car class.
This makes the epistemic uncertainty a measure for underrepresented classes.

Compared to the two-stage detectors used by \cite{di_feng_first} and \cite{wirges2019capturing} one-stage object detection yields uncertainty measures for the whole image and not just the boxes sampled by the RPN. This can be visually examined in \cref{fig:title}.
Incorporating uncertainty estimation into one-stage models has been done before. Le et al. \cite{Le_sampling_uncertainty} incorporate aleatoric loss attenuation as proposed by \cite{kendall_what} into the SSD one-stage detector \cite{Liu2016ssd}. As another measure of uncertainty they used the fact that SSD spawns many boxes. They grouped all boxes with a specific overlap and used the grouped boxes to calculated an uncertainty measure, where the uncertainty is high if the grouped boxes disagree on their predictions.
Miller et al. \cite{miller2018sampling} implemented a SSD architecture with dropout during test time. They employed a merging strategy of boxes from multiple forward passes based on their overlap. An uncertainty measure was calculated based on the merged boxes, effectively building a model ensemble, which is commonly used for uncertainty estimation.
The novelty of this article is that we apply the methods of \cite{Gal2016Bayesian} to use dropout as an approximation technique for the predictive distribution of a Bayesian neural network (BNN).

%% file: method.tex
\section{Method}
To generate uncertainty estimations we incorporate Bayesian deep learning into the YOLOv3 one-stage object detection framework.

\subsection{Bayesian Deep Learning}
\label{sec:bayesian-deep-learning}
Bayesian Neural Networks (BNNs) are neural networks where each weight is a random variable instead of a fixed value. A prior distribution $p(\omega)$ is placed over the networks weights $\omega$.
Together with a likelihood function $p(y|x,\omega)$ a posterior distribution over the networks weights is obtained by applying Bayes' theorem
\[
	p(\w|X, Y) = \frac{P(Y|X, \w)p(\w)}{P(Y|X)},
\]
with the marginal likelihood
\[
	p(Y|X) = \int p(Y|X, \w) p(\w) d\w,
\]
and $X, Y$ being the samples and ground truth of a given dataset. For regression tasks a Gaussian likelihood is used and for classification tasks a softmax likelihood.

For a new input $x^*$ we have the predictive distribution
\[
	p(y^*|x^*, X, Y) = \int p(y^*|x^*, \w)p(\w|X, Y)d\omega.
\]
In all but the simplest models the posterior and the predictive distribution are intractable.

One way to overcome this problem is to use variational inference where the posterior $p(\omega|X, Y)$ is approximated by a variational distribution $q_\theta(\omega)$ parametrised by $\theta$. This distribution can be found by minimizing the Kullback-–Leibler divergence $KL(q_\theta(\w)||p(\w|X,Y))$ or equivalently maximizing the \emph{evidence lower bound} (ELBO):
\begin{multline}
\label{eq:elbo}
    \mathcal{L}_{VI}(\theta) := \\
    \int q_\theta(\omega)\log p(Y|X, \omega) d\omega - \KL( q_\theta(\omega)||p(\omega)) d\w
\end{multline}
This approach results in an approximated variational predictive distribution
\begin{equation}
\label{eq:appr-predictive}
    q_\theta(y^*|x^*) = \int p(y^*|x^*, \w) q_\theta(\w)d\w.
\end{equation}

In \cite{Gal2016Bayesian} they showed that using dropout in neural networks can be interpreted as variational inference. They interpreted dropout to perform on the weights rather than on the outputs of a layer. This means performing dropout can be seen as sampling weights from the variational distribution $q_\theta(\w)$, effectively interpreting the  trained weights of the network as the parameters $\theta$ of the variational distribution. Weights with no dropout layer are viewed as delta distributed with the value of the weight being the mean.

Gal showed that training a neural network with dropout and l2 weight decay is equivalent to maximizing the ELBO. The l2 weight decay loss is important since minimizing it corresponds with minimizing the $\KL$ term in \eqref{eq:elbo}.
It is important to note that not only dropout can be seen as variational inference but all stochastic regularisation techniques which operate on the weights. The choice of regularisation technique manifests itself in an implicit prior distribution $p(\w)$. For dropout minimizing the l2 weight decay loss is equivalent to minimizing the $\KL$ term between $q_\theta(\w)$ and a normal distributed prior $p(\w)$.

If the models likelihood function is a (multivariate) normal distribution $p(y^* | f^\w(x^*)) = \mathcal{N}(y^*;  f^\w(x^*), \Sigma)$ Gal \cite{Gal2016thesis} showed that the expected value of the variational predictive distribution \eqref{eq:appr-predictive} can be approximated by Monte Carlo integration. This is done by sampling $T$ sets of weights $\tilde{\w}_t$ ($t = 1,\dots,T$) from the variational (dropout) distribution:
\begin{equation}
\label{eq:reg-predictive}
    \E_{q_\theta(y^*|x^*)} [y^*] \approx \frac{1}{T} \sum\nolimits_{t} f^{\tilde{\w}_t}(x^*)
\end{equation}
where $f^{\tilde{\w}_t}(x^*)$ is a forward pass through the net using the sampled weights $\tilde{\w}_t$. Essentially, performing $T$ forward passes through the network $f$ with dropout enabled. They called this process of summing up multiple forward passes with dropout enabled Monte Carlo (MC) dropout.
The variance of the variational predictive distribution can be obtained by:
\begin{multline*}
    \Var_{q_\theta(y^*|x^*)} [y^*] \\
    \approx
    \Sigma + \frac{1}{T} \sum\nolimits_{t} f^{\tilde{\w}_t}(x^*)^T f^{\tilde{\w}_t}(x^*) \\
    - \left(\sum\nolimits_{t} f^{\tilde{\w}_t}(x^*)\right)^T\left(\sum\nolimits_{t} f^{\tilde{\w}_t}(x^*)\right)
\end{multline*}

For classification tasks with either softmax (multiclass) or logistic (binary) likelihood  $p(y^*|x^*, \w) = s(f^\w(x^*))$ the variational predictive distribution can be approximated by:
\begin{equation}
\label{eq:cls-predictive}
    q_\theta(y^*|x^*) \approx \frac{1}{T} \sum\nolimits_{t} s(f^{\tilde{\w}_t}(x^*))
\end{equation}
As an uncertainty measure for classification tasks Gal \cite{Gal2016thesis} proposed the mutual information (MI):
\begin{multline}
\label{eq:mi}
    \I[y, \w|x, \D] := \\
    \HH[y|x, \D] - \E_{p(\w|\D)}\left[\HH[y|x,\w]\right]
    % \approx \\
    % - \sum_c\left(\frac{1}{T}\sum_t p(y=c|x, \tilde{w}_t)\right)
    % \log \left(\frac{1}{T}\sum_t p(y=c|x, \tilde{w}_t)\right)
    % \\+ \frac{1}{T}\sum_{t,c} p(y=c|x, \tilde{w}_t) \log p(y=c|x, \tilde{w}_t)
\end{multline}
with $\HH$ being the Shannon entropy.
% \begin{multline*}
%     \HH[y|x, \D] := \\
%     - \sum_c p(y=c|x,\D) \log  p(y=c|x,\D).
% \end{multline*}
It can also be approximated by sampling multiple dropout forward passes. Consult \cite{Gal2016thesis} for more information.
The mutual information is high for data where the model is uncertain on average but simultaneously, some weight configurations produced by the variational distribution yield outputs with high confidence.

\subsection{One-Stage Object Detection with YOLOv3}
YOLOv3 \cite{redmonYOLOv3} is a one-stage object detector which predicts boxes at three different scales.
The original input image is downsampled by a factor of 32, 16, and 8 respectively for the different output feature maps. For an input image of size $1024$ by $1920$ (height, width) this results in $32 \times 60$, $64 \times 120$, and $128 \times 240$ grids. Each cell of each grid encodes three different bounding boxes for three different prior box sizes (anchor boxes). Each bounding box consists of $4$ bounding box offsets, $1$ objectness score, and $c$ class scores. The objectness score signals if there is an object present for a given prior and cell.
Therefore, the three output grids are tensors of size $h_i \times w_i \times (3\cdot(4 + 1 + c))$  with $h_i$ and $w_i$ being height and width of the $i$-th output grid ($i = 1, 2, 3$).
In total, YOLOv3 produces $120\,960$ boxes for an input image of size $1024$ by $1920$. 
The bounding boxes from the raw output are calculated as follows. Let $c_x$ and $c_y$ denote the distance of the upper left corner of a cell to the upper left corner of the input image and $p_w$, $p_h$ denote the prior height and width of a given bounding box, then
\begin{align}
\begin{split}
\label{eq:box_encoding}
b_x =& \sigma(y^*_x) + c_x,\\
b_y =& \sigma(y^*_y) + c_y,\\
b_w =& p_w\exp(y^*_w),\\
b_h =& p_h\exp(y^*_h).
\end{split}
\end{align}
With the sigmoid (logistic) function $\sigma(x) := \frac{1}{1 + \exp(-x)}$ and the model output $y^* = (y^*_x, y^*_y, y^*_w, y^*_h, y^*_{obj}, y^*_{c_1}, \dots, y^*_{c_n})$ consisting of $4 + 1 + c$ entries.
The $x$ and $y$ coordinates of the bounding box center are denoted by $b_x$ and $b_y$, width and height by $b_w$ and $b_h$. The final objectness and class scores are calculated by applying the sigmoid activation function.

For calculating the loss each ground truth annotation (bounding box plus class label) is assigned to a single cell and prior (each cell holds 3 priors). Only the prior with the highest Intersection over Union (IoU) with the ground truth box receives a loss for the ground truth box (objectness, class and bounding box regression). After all ground truth boxes have been assigned, the remaining output cells and priors will receive only an objectness score loss to predict "no object".

In the following by prior we refer to the prior box sizes and not to the prior distribution. 

\subsection{Introducing Uncertainty to YOLOv3}
\label{sec:uncertainty-yolo}
We use a modified version of YOLOv3 which was implemented from scratch in TensorFlow\footnote{\url{https://github.com/flkraus/bayesian-yolov3}.}. The network architecture and ideas are the same as in the original implementation, however some changes were made.
The loss was carefully revised to fit into the Bayesian setting. A minor change was made regarding the class scores. The sigmoid activation was replaced with a softmax activation. We made this change since in our experiments all classes are mutually exclusive.
To predict aleatoric uncertainty for the bounding box regression we add 4 more entries $(y^*_{\sigma_x}, y^*_{\sigma_y}, y^*_{\sigma_w}, y^*_{\sigma_h})$ to the output per bounding box.

To fit our model to the methods of \cref{sec:bayesian-deep-learning} we need to model the output of our network as likelihood functions.

The bounding box regression is modelled as a multivariate normal distribution:
\[
    p(y^*\,|f^\w_{loc}(x^*)) = \mathcal{N}(y^*; f^\w_{loc}(x^*), \Sigma_{loc}),
\]
with mean $f^\w_{loc}(x^*)$ and diagonal covariance $\Sigma_{loc} = diag(\sigma_x^2, \sigma_y^2, \sigma_w^2, \sigma_h^2)$, where  $\Sigma_{loc}$ models the aleatoric uncertainty. Following \cite{kendall_what}, our aleatoric models include  $(\sigma_x^2, \sigma_y^2, \sigma_w^2, \sigma_h^2)$ as an explicit output of the network. For the models without aleatoric uncertainty we set them to $1$ during training and test time.

We use the negative log-likelihood loss for the localization:
\begin{multline*}
    \mathcal{L}(y_{loc}, f^\w_{loc}(x^*)) = 
    -\log p(y_{loc}|f^\w_{loc}(x^*)) = \\
    \sum\nolimits_i \left(
        \frac{\sigma_i^{-2}}{2} \left(y_i - f^\w_{i}(x^*)\right)^2 
        + \frac{1}{2}\log \sigma_i^{2}
    \right)
    + 2\log 2\pi
\end{multline*}
with $i = x,y,w,h$.

Note, that we model the raw outputs $y^*$ to follow a multivariate normal distribution and not the final bounding box coordinates $b_x$, $b_y$, $b_w$, and $b_h$. This also means we have to invert eqs. \eqref{eq:box_encoding} for our ground truth boxes to get to $y_i$\footnote{The inverting of the bounding box formulas for the ground truth is also mentioned in the original YOLOv3 paper but is not present in the code. Instead the loss is $\sigma(y^*_x) - y_x$. Also, their use of the l1 loss instead of an l2 loss does not allow us to view this as the negative log-likelihood of a normal distribution.}.

Following \cite{kendall_what} instead of predicting $\sigma^2_i$ we predict ${s_i := \log \sigma^2_i}$ to improve numerical stability:
\begin{multline*}
\mathcal{L}(y_{loc}, f^\w_{loc}(x^*)) = \\
    \sum\nolimits_i \left(
        \frac{1}{2}\exp(-s_i) \left(y_i - f^\w_{i}(x^*)\right)^2 
        + \frac{1}{2} s_i
    \right),
\end{multline*}
where we also ditched the constant term. To avoid NaN errors during training we also clip $s_i$ to the interval $[-40, 40]$, this is reasonable since $\exp(\pm 40) \approx \num{1e\pm17}$ covers a wide enough range. When setting $\sigma^2_i = 1$ (or $s_i = 0$) this loss is equivalent to the standard l2 regression loss $\frac{1}{2}\norm{y_{loc} - f^\w_{loc}(x^*)}^2$. We call those predicted variances $\sigma_i^2$ aleatoric variances or uncertainty.

The objectness score is modeled as a logistic likelihood:
\begin{multline*}
    p(y^* = c \,| f^\w_{obj}(x^*)) = \\
    c \cdot \sigma(f^\w_{obj}(x^*)) + (1 - c)(1 - \sigma(f^\w_{obj}(x^*)))
\end{multline*}
with $c \in \{0,1\}$, representing the presence or absence of an object. For the loss we use the negative log-likelihood, which, for the sigmoid likelihood, is the same as the cross entropy loss:
\begin{multline*}
    \mathcal{L}(y_{obj}, f^\w_{obj}(x^*)) = \\
    - y_{obj}\cdot \log \sigma(f^\w_{obj}(x^*)) \\
    - (1 - y_{obj})\cdot \log (1 - \sigma(f^\w_{obj}(x^*))),
\end{multline*}
where $y_{obj} \in \{0, 1\}$ is the ground truth label.

The classification part is modeled as a softmax likelihood:
\[
    p(y^*|f^\w_{cls}(x^*)) = \text{softmax}(f^\w_{cls}(x^*)),
\]
again with the standard cross entropy loss (negative log-likelihood). Modelling all our losses as negative log likelihood allows us to use the techniques proposed by \cite{Gal2016Bayesian}.

We combine all three losses into a single likelihood function by taking the product of the three individual likelihood functions:
\begin{multline*}
    p(y^*|f^\w(x^*)) = \\
        p(y^*_{loc}| f^\w_{loc}(x^*)) \cdot p(y^*_{obj}| f^\w_{obj}(x^*)) \cdot  p(y^*_{cls}| f^\w_{cls}(x^*))
\end{multline*}
The combined loss is the sum of the three individual losses which is the same as the negative log-likelihood of the combined likelihood function
\begin{multline*}
    \mathcal{L}(y, f^\w(x^*)) = 
    \mathcal{L}(y_{loc}, f^\w_{loc}(x^*)) \\
    + \mathcal{L}(y_{obj}, f^\w_{obj}(x^*)) + \mathcal{L}(y_{cls}, f^\w_{cls}(x^*)).
\end{multline*}
We also add a l2 weight decay $\mathcal{L}_{l2}(\w) = \frac{1}{2} \lambda \norm{w}^2$ to the loss. In doing so minimizing our loss becomes equivalent to maximizing the ELBO \eqref{eq:elbo}.
\input{iou_plots.tex}
\subsection{Bayesian YOLO}
We use the same network architecture as YOLOv3 which uses darknet53 \cite{redmonYOLOv3} as a feature extractor (base net) and on top of that several convolutional layers mixed with upsampling layers to achieve the different downsampling scales of $32$, $16$, and $8$. All layers are convolutional layers making it invariant to the input size, only the prior sizes must be adjusted accordingly. We want to emphasize that our method is agnostic to the network architecture itself as long as the here presented loss is used.

To predict epistemic uncertainty we add dropout to our model. Before each output feature map we add dropout to five convolutional layers. In total 15 dropout layers are added.
The dropout is applied right after the convolutional layer and before the batch normalization layer. Note that we did not remove the batch normalization layer. Batch normalization can be seen as another non linear activation and does not interfere with the variational inference as long as we place the dropout layer right after the convolutional layer. We used a dropout rate of $0.1$. We also experimented with a dropout rate of $0.2$ but this yielded worse model performance.
Leaving in the batch normalization layers, stabilizes the training and we can easily use weights of a pre-trained model without dropout to jump-start the training of the model with dropout.

Our model predicts boxes at three different scales and for nine different prior sizes. This can be interpreted as a combined model consisting of nine different (sub-)models which share most weights. Each of these sub-models predicts $n_i$ boxes ($i \in \{1,\dots,9\}$), with $n_i$ depending on the input image size. Each of these boxes must be interpreted as the output of the likelihood functions introduced in \cref{sec:uncertainty-yolo}. For each box the regression parameters are calculated by approximating the expected value of the variational predictive distribution by applying \eqref{eq:reg-predictive} and the class and objectness scores by approximating the predictive distribution via \eqref{eq:cls-predictive}. For our model with aleatoric loss and MC dropout enabled the aleatoric variance is calculated by averaging out multiple forward passes.
For this process to make sense we utilize the fact that the box predictions of YOLOv3 are locally restricted to the cell they are predicted by. Therefore multiple forward passes of the same box with dropout enabled still refer to the same location in the input image.

This is different to SSD where the bounding box offsets are not restricted to the enclosing cell but can, in theory, shift a box over the whole image. This causes problems if multiple dropout forward passes of one output predict offsets for different objects. This might be the reason why \cite{miller2018sampling} grouped boxes of multiple forward passes based on their overlap rather than averaging out the network output to approximate the variational predictive distribution.

The nine in one model of YOLOv3 is quite different to the two-stage approach taken by \cite{di_feng_first}. The last fully connected layers which they used to predict the uncertainties is the same for all region proposals. The effect is that the uncertainty estimations come all from the same model with the same weights which is not the case for our Bayesian YOLO.

\subsection{Training and Dataset}
\label{sec:training-and-dataset}
All models are first trained without dropout and aleatoric loss for $125\,000$ iterations. Each iteration is the forward and backward pass of one batch. We used a batch size of $3$ for all models.
We then use the checkpoint after $125\,000$ iterations to train three models. One model with dropout enabled (as described above), one without dropout but the aleatoric loss enabled and one combined model with dropout and aleatoric loss enabled (referred to as aleatoric+epistemic).
The initial training without dropout and aleatoric loss is crucial for stable training as found by \cite{wirges2019capturing} and confirmed by us.

We continued the training for the model without any uncertainty as a baseline. All models were trained for further $575\,000$ iterations to a total of $700\,000$ with the first $125\,000$ iterations shared between all four models.
For all experiments we used an Adam optimizer \cite{Kingma2015Adam} with an initial learning rate of \num{e-5}.
The prior sizes are determined by applying the k-means clustering to the box sizes of the training dataset as described in \cite{Redmon_2017_CVPR}.

For our experiments we use the EuroCity Persons (ECP) dataset \cite{braun2019ECP} which consists of roughly 47000 images of which around 7000 are captured at night. In total around 218000 pedestrians and 20000 riders are labeled\footnote{The training and validation splits including annotations are available to the public at \url{https://eurocity-dataset.tudelft.nl}.}. The dataset provides annotations for the level of occlusion of every object. The labels have the following granularity: no occlusion, $11\%-40\%$, $41\%-80\%$, and $>80\%$. We use these occlusion labels in our experiments.

%% file: iou_plots.tex
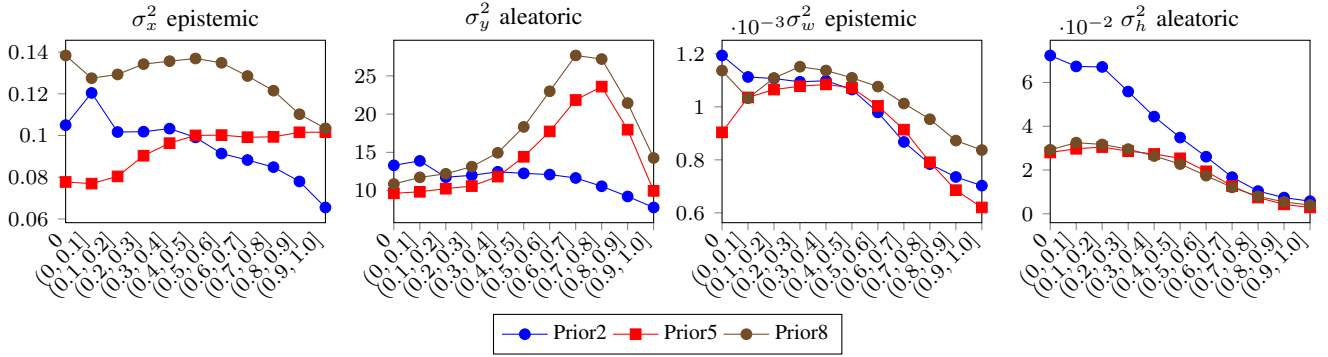
\begin{figure*}[ht]
\centering
\begin{tikzpicture}
\begin{groupplot}[
group style={
group name=my plots,
group size=4 by 1,
ylabels at=edge left,
vertical sep=1.5cm,
horizontal sep=0.9cm,
},
footnotesize,
width=5cm,
height=4cm,
tickpos=left,
ytick align=outside,
xtick align=outside,
enlarge x limits=false,
xticklabels={0, $(0{,}\, 0.1]$, $(0.1{,}\,0.2]$, $(0.2{,}\,0.3]$, $(0.3{,}\,0.4]$, $(0.4{,}\,0.5]$, $(0.5{,}\,0.6]$, $(0.6{,}\,0.7]$, $(0.7{,}\,0.8]$, $(0.8{,}\,0.9]$, $(0.9{,}\,1.0]$},
xtick={1,...,11},
cycle list name=color,
xticklabel style = {rotate=45,anchor=east},
yticklabel style={
        /pgf/number format/fixed,
        /pgf/number format/precision=2
},
legend style={at={(-0.8,-0.5)}, legend columns=3},
]
\nextgroupplot[title={$\sigma^2_x$ epistemic}, xtick style={draw=none}]
\addplot
coordinates {
% key: x_var_epi, prior: 1
(1, 0.10497628527073972)(2, 0.12042353333616838)(3, 0.10168656952604889)(4, 0.10184713804486409)(5, 0.10330760847897535)(6, 0.09911604551360563)(7, 0.09136474276532065)(8, 0.08826866545545091)(9, 0.08485566136165401)(10, 0.07799450237187468)(11, 0.0655226324849269)
};

\addplot
coordinates {
% key: x_var_epi, prior: 4
(1, 0.07774002984215865)(2, 0.07695630749560449)(3, 0.08041317427269823)(4, 0.09033898433488911)(5, 0.09634117789815723)(6, 0.10006465725264445)(7, 0.1001901115101584)(8, 0.09920052385415574)(9, 0.09937682574366988)(10, 0.10153722851513855)(11, 0.10161861855419674)
};

\addplot
coordinates {
% key: x_var_epi, prior: 7
(1, 0.1383826244235797)(2, 0.12748642191796114)(3, 0.12928774136599536)(4, 0.13423357423011964)(5, 0.13564891216240882)(6, 0.13692345569221107)(7, 0.13486869731192144)(8, 0.12851736455711932)(9, 0.12150693694821547)(10, 0.1102178169443611)(11, 0.10344597941150828)
};

\nextgroupplot[title={$\sigma^2_y$ aleatoric}]

\addplot
coordinates {
% key: y_var_ale, prior: 1
(1, 13.298725909582862)(2, 13.884068939115236)(3, 11.743955271404282)(4, 11.992938451823091)(5, 12.452721279816796)(6, 12.253401022710925)(7, 12.09077885337531)(8, 11.649503655541006)(9, 10.562622392896817)(10, 9.231011406144143)(11, 7.791389482161578)
};

\addplot
coordinates {
% key: y_var_ale, prior: 4
(1, 9.642532369204178)(2, 9.841767296600665)(3, 10.246645931514598)(4, 10.58759479039585)(5, 11.82626032612437)(6, 14.421094308916054)(7, 17.737041662597)(8, 21.842356748691007)(9, 23.594970468149157)(10, 17.96322165596757)(11, 9.96515660984978)
};

\addplot
coordinates {
% key: y_var_ale, prior: 7
(1, 10.85392728663234)(2, 11.72980506457334)(3, 12.191764256774034)(4, 13.128367691915043)(5, 14.937469819353899)(6, 18.329259849196188)(7, 22.989407198555718)(8, 27.667591819079043)(9, 27.195269451564464)(10, 21.442440375418318)(11, 14.262143129697987)
};

\nextgroupplot[title={$\sigma^2_w$ epistemic},]

\addplot
coordinates {
% key: w_var_epi, prior: 1
(1, 0.001193772839607974)(2, 0.001112694717272514)(3, 0.0011067380779789988)(4, 0.001094458533635729)(5, 0.0010983877522632419)(6, 0.0010652887032637749)(7, 0.0009790065499371973)(8, 0.0008675813516325722)(9, 0.0007840325857784974)(10, 0.0007353555563057459)(11, 0.0007027004557826063)
};

\addplot
coordinates {
% key: w_var_epi, prior: 4
(1, 0.0009038672164364416)(2, 0.0010355213839134047)(3, 0.00106509139423779)(4, 0.001077736091429839)(5, 0.0010844604986881264)(6, 0.0010715848153042664)(7, 0.0010034994647482981)(8, 0.0009140924242822663)(9, 0.0007909291202058369)(10, 0.0006856785307709107)(11, 0.000620422047896296)
};

\addplot
coordinates {
% key: w_var_epi, prior: 7
(1, 0.0011365817739552217)(2, 0.0010324559758590037)(3, 0.0011087854770261364)(4, 0.0011513835361764834)(5, 0.0011370829234884866)(6, 0.0011095266635419527)(7, 0.0010763842714656793)(8, 0.0010126309939102079)(9, 0.0009532887965496848)(10, 0.0008727405713846497)(11, 0.0008369523499537032)
};

\nextgroupplot[title={$\sigma^2_h$ aleatoric}, name=h-var-ale]

\addplot
coordinates {
% key: h_var_ale, prior: 1
(1, 0.07233512741151037)(2, 0.06732805268412244)(3, 0.06709602421641202)(4, 0.05584344668996859)(5, 0.0444128639941704)(6, 0.03482543922574706)(7, 0.026089926860405576)(8, 0.016746140894579865)(9, 0.01037328947069176)(10, 0.007391588540582951)(11, 0.005752103021094466)
};

\addplot
coordinates {
% key: h_var_ale, prior: 4
(1, 0.02801013207809163)(2, 0.02970371848731586)(3, 0.030488837561425685)(4, 0.028636849463324753)(5, 0.027331166565845798)(6, 0.025334875550578062)(7, 0.01940913275793726)(8, 0.012665042250721315)(9, 0.0074023861964186595)(10, 0.004306667474186662)(11, 0.0028883054409376027)
};

\addplot
coordinates {
% key: h_var_ale, prior: 7
(1, 0.029245329949896627)(2, 0.03249671374287259)(3, 0.031681337758920396)(4, 0.029550415505287638)(5, 0.026391648218620996)(6, 0.022731600153055743)(7, 0.017430298482249862)(8, 0.012048373479012772)(9, 0.007983472286747994)(10, 0.005409094497480278)(11, 0.0039878899027143075)
};

\legend{Prior2, Prior5, Prior8}

\end{groupplot}

\end{tikzpicture}

\caption{Uncertainties compared to intersection over union (IoU), we show one prior size from each output scale. The x-axis shows the IoU ranges.}
\label{fig:iou-ucty-plot}
\end{figure*}

%% file: experiments.tex
\section{Experiments and Results}
\label{sec:experiments}

\begin{table}[t]
\centering
\begin{tabular}{ccc}
	\toprule
	    T      &      LAMR      &  time (ms)   \\ \midrule
	no dropout &    $10.91$     & \textbf{120} \\
	    5      &    $10.38$     &    $203$     \\
	    10     &    $10.40$     &    $257$     \\
	    20     &    $10.35$     &    $411$     \\
	    40     &    $10.35$     &    $743$     \\
	    50     & \textbf{10.33} &    $912$     \\
	   100     &    $10.34$     &    $1907$    \\ \bottomrule
\end{tabular}
\caption{Performance difference for different number of stochastic forward passes and inference time for full image size of $1024 \times 1920$.}
\label{tab:mcdrop}
\end{table}

\begin{table}[ht]
\centering
\begin{tabular}{@{}ccccc@{}}
	\toprule
	                                &                       &    \multicolumn{2}{c}{LAMR}     &           \\
	\cmidrule(r){3-4}
	         Training Data          &         Model         &      Day       &     Night      &   Steps   \\ \midrule
	              day               &       baseline        &    $11.00$     &    $24.23$     & $ 550000$ \\
	              day               &       aleatoric       &    $11.20$     &    $25.60$     & $ 650000$ \\
	              day               &       epistemic       & \textbf{10.65} &    $25.71$     & $ 625000$ \\
	              day               & aleatoric + epistemic &    $10.70$     & \textbf{22.68} & $ 600000$ \\ \midrule
	             night              &       baseline        &    $28.00$     &    $22.34$     & $ 525000$ \\
	             night              &       aleatoric       &    $28.45$     &    $21.70$     & $ 175000$ \\
	             night              &       epistemic       & \textbf{25.64} &    $21.12$     & $ 700000$ \\
	             night              & aleatoric + epistemic &    $28.44$     & \textbf{20.74} & $ 700000$ \\ \midrule
	          day + night           &       baseline        &    $11.10$     &    $15.06$     & $ 625000$ \\
	          day + night           &       aleatoric       &    $10.42$     & \textbf{13.38} & $ 300000$ \\
	          day + night           &       epistemic       &    $10.75$     &    $15.08$     & $ 625000$ \\
	          day + night           & aleatoric + epistemic & \textbf{10.33} &    $14.52$     & $ 575000$ \\ \bottomrule
\end{tabular}
\caption{Model Performance on ECP test set, measured with LAMR metric (lower is better).}
\label{tab:performance}
\end{table}

\input{iou_mi_plots.tex}

To train our models we split the training data into day, night and day + night. Each split was used to train 4 models as described in \cref{sec:training-and-dataset}.
The models were trained to predict pedestrians and riders. We applied standard non-maximum suppression (TensorFlow implementation) for each class individually with an IoU-threshold of $0.5$.

To measure model performance we use the evaluation scripts provided by \cite{braun2019ECP}. The performance metric used is the \emph{log average miss-rate} (LAMR), lower scores in this metric are better. We only evaluate for pedestrians using the \emph{reasonable} difficulty and \emph{ignore} setting of the evaluation script. These settings are such that boxes which are too small don't affect the score and riders detected as pedestrians are ignored. These same settings are used by \cite{braun2019ECP}.

First, we selected the best performing checkpoint of each model on the validation set. For each model we evaluated checkpoints every $25\,000$ iterations.
The models trained on the day or day + night data were evaluated on the day validation set and the models trained only on night data on the night validation set.
Then, we evaluated the best performing model-checkpoints on the day and night test set. To evaluate the epistemic models we used MC dropout, with $T=50$ forward passes per image.

\subsection{Performance Evaluation}
In \cref{tab:performance} the performance of the different models are compared.
We found that employing the aleatoric loss and dropout slightly increased model performance when training on day and night data. When training on night data alone both the aleatoric loss and dropout increased the performance on the night validation set but got slightly worse on the day validation set compared to the baseline. Overall it seems that both methods slightly increase the performance over the baseline while providing additional uncertainty estimates. 

In \cref{tab:mcdrop} we evaluate the effect of different numbers of forward passes for the MC dropout. We use our best performing model \emph{aleatoric + epistemic} trained on day and night data. The model is evaluated on the day test set. Performance and inference time are presented for different $T$ and for the standard dropout approximation with only one forward pass. As expected the number of forward passes slightly increases model performance. However for more than 20 forward passes the difference is minute.
An interesting fact from this evaluation is, that the standard dropout procedure did not perform significantly worse, and slightly outperforms the baseline approach (compare \cref{tab:mcdrop}).

To save time the input images are only passed once through the base feature extractor, since no dropout is applied during this stage of the network. The output of the feature extractor is stacked $T$ times to a single batch with $T$ identical entries and then processed in parallel by the subsequent layers. Although done in parallel this still drastically impacts performance. However, for a full sized image of 1024 by 1920 going from one forward pass to ten doubles the inference time from $120$ms to $257$ms, which, with further optimizations applied, could still be manageable.

\subsection{Uncertainty Estimations}
\begin{figure}
\centering

\begin{subfigure}{.46\columnwidth}
\includegraphics[width=\columnwidth]{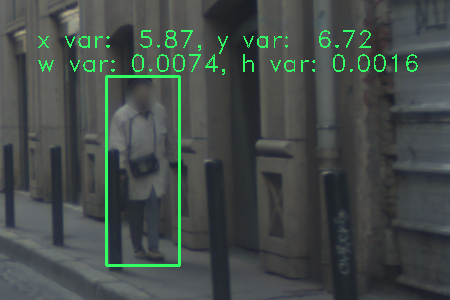}
\caption{Prediction for fully visible pedestrian.}
\end{subfigure}
\begin{subfigure}{.46\columnwidth}
\includegraphics[width=\columnwidth]{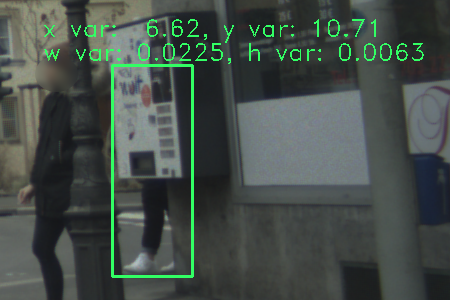}
\caption{Prediction for heavily occluded pedestrian.}
\end{subfigure}
\caption{Comparing aleatoric uncertainties for occluded and fully visible pedestrians. Both Predictions were produced by prior 3. Note the drastically increased variance in width and height for the heavily occluded pedestrian.}
\label{fig:ale-occlusion}

\end{figure}

\input{qa.tex}

When evaluating the uncertainty measures we find that the uncertainties between the individual priors are not calibrated, this can be seen in \cref{fig:iou-ucty-plot}, where the average uncertainty for prior $2$ clearly differs from prior $8$. As a result we can only compare the uncertainties for each prior individually. This is different from the uncertainty estimation of \cite{di_feng_first}, where they used a two-stage object detector. The two-stage detector has a single output to predict the class and box refinements of the region proposal network. Making the uncertainty estimation comparable between different boxes.

Since both YOLOv3 and SSD predict boxes on multiple feature maps, using uncertainty estimations for these type of networks requires additional work to calibrate the outputs. Investigation of suitable calibration methods will be reserved for future research.

When comparing the variance for the regression task we found for both the aleatoric and the epistemic uncertainty that the variance for the bounding box offset was higher than for the width and height adjustment. This seems plausible, since the prior sizes were carefully chosen to fit the training data well, so less adjustment is needed.

We carried out a qualitative analysis of the uncertainty estimates. We overlaid images with the uncertainty estimations for each cell of each prior, see \cref{fig:qa}. For the regression uncertainties, both aleatoric and epistemic, it is clearly visible that the center cell of each object has the lowest uncertainty. The cells around them have higher uncertainty which decreases further away, with large parts of the background having low uncertainty (see \cref{fig:title}).

The variance for the $x$ and $y$ offset predictions are particularly interesting. For the $x$ variance we found multiple cases where the uncertainty is low in a center strip and high to the left and right. Conversely for the $y$ variance we find the uncertainty to be high above and below the center (see \cref{fig:y-ale-center-certain,fig:x-ale-center-certain}).
This phenomenon happens for both epistemic and aleatoric uncertainty estimations. Although it seems that it is more pronounced for the aleatoric uncertainty.

The uncertainty estimations for the height and width do not show such clear patterns, although we still find the center to have the lowest uncertainty. Altogether the epistemic variance seems more spread out compared to the aleatoric.

Having low uncertainty estimations for the cell which is responsible for predicting an object is what we would expect from a sensible uncertainty estimations. We want to recall, that in the YOLO framework only one cell is responsible for predicting an object. This differs from the SSD framework where multiple cells are trained to predict a single ground truth object.
On the other hand it is import to have higher uncertainty for cells which are slightly off, indicating that the bounding box is not quite right.
And we do not care as much about the uncertainty estimations for the background.

\begin{table*}[ht]
\centering
\begin{tabular}{@{}ccccccccccc@{}}
	\toprule
	             &     &  Prior1  &  Prior2  &  Prior3  &  Prior4  &  Prior5  &  Prior6  &  Prior7  &  Prior8  &  Prior9  \\ \midrule
	$\sigma^2_w$ & $r$ & $0.070$  & $0.359$  & $0.490$  & $0.499$  & $0.474$  & $0.466$  & $0.392$  & $0.254$  & $0.025$  \\
	             & $p$ & $0.374$  & $<0.000$ & $<0.000$ & $<0.000$ & $<0.000$ & $<0.000$ & $<0.000$ & $<0.000$ & $0.118$  \\ \midrule
	$\sigma^2_h$ & $r$ & $-0.005$ & $0.214$  & $0.383$  & $0.202$  & $0.239$  & $0.349$  & $0.445$  & $0.371$  & $0.140$  \\
	             & $p$ & $0.947$  & $<0.000$ & $<0.000$ & $<0.000$ & $<0.000$ & $<0.000$ & $<0.000$ & $<0.000$ & $<0.000$ \\ \midrule
	 data count  &     &  $164$   &  $408$   &  $828$   &  $1821$  &  $2668$  &  $3252$  &  $5635$  &  $5825$  &  $3936$  \\ \bottomrule
\end{tabular}
\caption{Pearson correlation coefficient $r$ between occlusion and aleatoric uncertainty for $w$ and $h$. The last row shows the number of detections used to calculate $r$ for each prior size. This was calculated using the epistemic + aleatoric model trained on day and night data. The occlusion was evaluated on the day validation set. We found the same tendencies for different models and when evaluating on the night validation set.}
\label{tab:ale-corr}
\end{table*}

The effect of having low uncertainties in the center can also be seen in the quantitative analysis carried out in \cref{fig:iou-ucty-plot}. We plotted the different uncertainty measures against the overlap (IoU) with a ground truth object. We clearly see the uncertainty going down for an overlap greater than $0.7$. The aleatoric variance for the $y$ offset even ramps up for detections which are slightly off with an overlap of around $0.8$ and then going down for more precise detections.

We also evaluated the Mutual Information (MI) (see eq. \eqref{eq:mi}) for the class and objectness score. The MI for the class score is mostly low but sometimes shows higher values around objects (see \cref{fig:cls-mi-around}). For some difficult objects, such as strollers, or pedestrians standing in front of a bicycle the MI goes up (see \cref{fig:cls-mi-stroller,fig:cls-mi-ped-infront-bicycle}). Since the model was trained on riders and pedestrians a person standing in front of a bicycle has both features for the pedestrian and the rider class increasing the MI for the class prediction.

The MI for the objectness score does not predict sensible uncertainty information when evaluating it. The uncertainty is very low for the complete background and only high on objects. With the center of the object having the highest uncertainty (see \cref{fig:obj-mi-center-high}). This can also be seen in \cref{plt:obj-mi-iou} where the MI correlates positively with the IoU with an ground truth object.

This behaviour is likely a result of the training process. The objectness score is the only output which is also trained on background. Since most parts of an image are background this causes the model to become very confident on the background. Unfortunately this makes the MI of the objectness not a practical measure of uncertainty, since correct predictions have high and incorrect ones low uncertainty.
Maybe this effect could be counteracted by training for more iterations or trying to balance out the background and foreground examples during training as SSD does.
The other uncertainty estimations do not suffer from such effects since the other model outputs only incur a loss for positive examples.

\subsection{Occlusion}
We evaluated how the occlusion of an ground truth object affects the uncertainty estimations.
We found the aleatoric uncertainty for the bounding box regression to be positively correlated with the occlusion of the detected object.

To evaluate this we selected for each ground truth object the model output which is responsible for predicting it (similar to the training process). We calculate the Pearson correlation coefficient between the occlusion level of the matched ground truth ($0\%$, $11\%-40\%$, $41\%-80\%$, $>80\%$). We find that the variance of the bounding box regression parameters for width and height have a strong positive correlation between the occlusion and the predicted variance, see \cref{tab:ale-corr}. This seems plausible since occluded objects in the ECP dataset have an estimated bounding box of the full extent. These, by humans estimated, boxes are used during the training phase forcing the model to guess the correct width and height of the object, possibly predicting high aleatoric variance to keep the loss low. We also found a slight correlation between the occlusion and aleatoric variance for the bounding box offsets. The epistemic uncertainty did not show any correlation with the occlusion.

A qualitative analysis of this can be found in \cref{fig:ale-occlusion} where we compared the aleatoric uncertainty of an occluded and a fully visible pedestrian. Both boxes are produced by prior~$3$ so the uncertainty measures are comparable. We found that when comparing the uncertainties of different priors an occluded box can have lower uncertainty than a non-occluded box from a different prior, which again, stresses the importance of keeping the uncertainty measures of different priors separate or calibrate them, to make them comparable.

%% file: iou_mi_plots.tex
\begin{figure}[ht]\centering\begin{tikzpicture}
\begin{axis}[
ylabel=Objectness MI,
xlabel=IoU,
x label style={at={(axis description cs:0.95, -0.1)},anchor=north, rotate=0},
xticklabels={0, $(0{,}\, 0.1]$, $(0.1{,}\,0.2]$, $(0.2{,}\,0.3]$, $(0.3{,}\,0.4]$, $(0.4{,}\,0.5]$, $(0.5{,}\,0.6]$, $(0.6{,}\,0.7]$, $(0.7{,}\,0.8]$, $(0.8{,}\,0.9]$, $(0.9{,}\,1.0]$},
xtick={1,...,11},
width=.9\columnwidth,
height=5.5cm,
cycle list name=black white,
legend pos=north west,
xticklabel style = {rotate=45,anchor=east},
legend style={legend columns=2},
]

\addplot
coordinates {
% key: obj_mutual_info, prior: 0
(1, 5.146157969081884e-06)(2, 3.0201564127525453e-06)(3, 1.3046534037299796e-05)(4, 1.571745296403952e-05)(5, 6.955408098888425e-05)(6, 0.000323922400493243)(7, 0.001155027226933864)(8, 0.002433246106146252)(9, 0.003966342304681471)(10, 0.004191973021376804)(11, 0.005864100645116547)
};

\addplot
coordinates {
% key: obj_mutual_info, prior: 1
(1, 9.160679479392924e-06)(2, 9.69720544893441e-06)(3, 3.347457819371057e-05)(4, 6.00095118346729e-05)(5, 0.00015729176958032812)(6, 0.00038113139431933705)(7, 0.0008217968958226839)(8, 0.0020548624032264416)(9, 0.005289268836683273)(10, 0.010355844322014961)(11, 0.015541223848156411)
};

\addplot
coordinates {
% key: obj_mutual_info, prior: 2
(1, 1.3250172409246481e-05)(2, 2.93101748465275e-05)(3, 4.7734884299047224e-05)(4, 8.905978806042258e-05)(5, 0.0001620996583392681)(6, 0.0002947566990845755)(7, 0.000649217580909191)(8, 0.0014373125627127733)(9, 0.0040852273295004255)(10, 0.010880312904736738)(11, 0.017937262018314182)
};

\addplot
coordinates {
% key: obj_mutual_info, prior: 3
(1, 1.7227012129174512e-05)(2, 3.990552861954993e-05)(3, 6.36404347635281e-05)(4, 0.00011089676119248637)(5, 0.0001963982981567818)(6, 0.00039812876435930527)(7, 0.0009463338807980898)(8, 0.002567762944753965)(9, 0.006849736727566169)(10, 0.014459184381981042)(11, 0.02039344771099942)
};

\addplot
coordinates {
% key: obj_mutual_info, prior: 4
(1, 4.367039723402482e-05)(2, 0.00011895987395864101)(3, 0.00016372441569472638)(4, 0.0002577669046853104)(5, 0.00039601958388575225)(6, 0.0006722717367018373)(7, 0.0011888803225498126)(8, 0.0024371131305182423)(9, 0.00607603859694912)(10, 0.014350024372114512)(11, 0.0208544430303441)
};

\addplot
coordinates {
% key: obj_mutual_info, prior: 5
(1, 4.7851071410613066e-05)(2, 0.00012360609378283128)(3, 0.00015722417864798917)(4, 0.00024234781454648486)(5, 0.0003737535539030945)(6, 0.000617635118375387)(7, 0.0012236509996639384)(8, 0.0027262797455593494)(9, 0.006344323208010672)(10, 0.013596900229569686)(11, 0.018367306096229746)
};

\addplot
coordinates {
% key: obj_mutual_info, prior: 6
(1, 3.1994670580505586e-05)(2, 6.037543796432577e-05)(3, 7.477475718387972e-05)(4, 0.00010654191200996159)(5, 0.0001754393196153257)(6, 0.0003537568066693171)(7, 0.0008008922100609989)(8, 0.0021143493778146357)(9, 0.0055404980678248625)(10, 0.012335675602715753)(11, 0.01851291807661099)
};

\addplot
coordinates {
% key: obj_mutual_info, prior: 7
(1, 4.698785107367116e-05)(2, 8.085076030317274e-05)(3, 0.00012145842494303024)(4, 0.0001762630808374607)(5, 0.0003129522852046887)(6, 0.0006084270782541885)(7, 0.001349400616979309)(8, 0.0030322989885593317)(9, 0.007013685316117905)(10, 0.012230089046683237)(11, 0.015623551259252369)
};

\addplot
coordinates {
% key: obj_mutual_info, prior: 8
(1, 7.097390743493694e-05)(2, 6.26568192170186e-05)(3, 0.00012184902742863391)(4, 0.0002362240442790684)(5, 0.0004881323078423607)(6, 0.0010124527211728032)(7, 0.002165751942282211)(8, 0.0046057390510020925)(9, 0.008846300498752215)(10, 0.012931108037495696)(11, 0.01569678543291225)
};

\legend {Prior1, Prior2, Prior3, Prior4, Prior5, Prior6, Prior7, Prior8, Prior9}
\end{axis}
\end{tikzpicture}
\caption{Mutual information of the objectness score plotted against the overlap with an ground truth object.}
\label{plt:obj-mi-iou}
\end{figure}

%% file: qa.tex
\begin{figure*}[ht]
\centering

\begin{subfigure}{.13\textwidth}
\includegraphics[width=\textwidth]{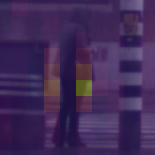}
\caption{{$\sigma_x^2$ aleatoric}}
\label{fig:x-ale-center-certain}
\end{subfigure}
\begin{subfigure}{.13\textwidth}
\includegraphics[width=\textwidth]{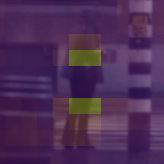}
\caption{{$\sigma_y^2$ aleatoric}}
\label{fig:y-ale-center-certain}
\end{subfigure}
\begin{subfigure}{.13\textwidth}
\includegraphics[width=\textwidth]{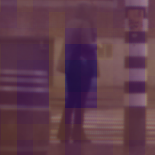}
\caption{{$\sigma_w^2$ aleatoric}}
\end{subfigure}
\begin{subfigure}{.13\textwidth}
\includegraphics[width=\textwidth]{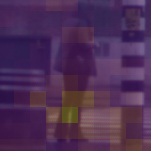}
\caption{{$\sigma_y^2$ epistemic}}
\end{subfigure}
\begin{subfigure}{.13\textwidth}
\includegraphics[width=\textwidth]{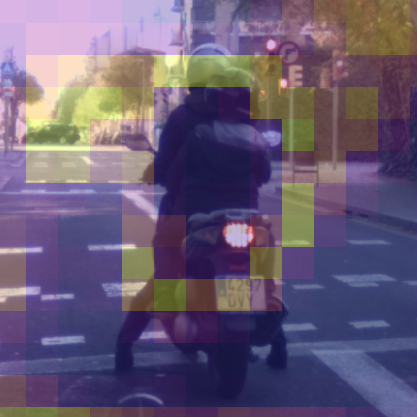}
\caption{Class MI}
\label{fig:cls-mi-around}
\end{subfigure}
\begin{subfigure}{.13\textwidth}
\includegraphics[width=\textwidth]{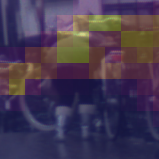}
\caption{Class MI}
\label{fig:cls-mi-ped-infront-bicycle}
\end{subfigure}
\begin{subfigure}{.13\textwidth}
\includegraphics[width=\textwidth]{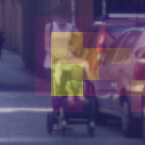}
\caption{Class MI}
\label{fig:cls-mi-stroller}
\end{subfigure}

\begin{subfigure}{.13\textwidth}
\includegraphics[width=\textwidth]{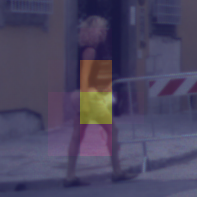}
\caption{Obj. MI}
\label{fig:obj-mi-center-high}
\end{subfigure}
\begin{subfigure}{.13\textwidth}
\includegraphics[width=\textwidth]{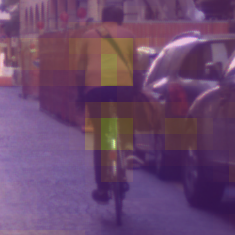}
\caption{{$\sigma_y^2$ aleatoric}}
\end{subfigure}
\begin{subfigure}{.13\textwidth}
\includegraphics[width=\textwidth]{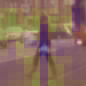}
\caption{{$\sigma_w^2$ aleatoric}}
\end{subfigure}
\begin{subfigure}{.13\textwidth}
\includegraphics[width=\textwidth]{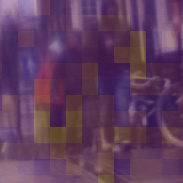}
\caption{{$\sigma_w^2$ epistemic}}
\end{subfigure}
\begin{subfigure}{.13\textwidth}
\includegraphics[width=\textwidth]{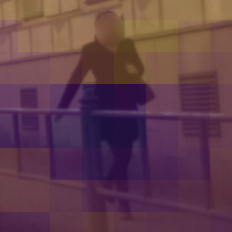}
\caption{{$\sigma_h^2$ aleatoric}}
\end{subfigure}
\begin{subfigure}{.13\textwidth}
\includegraphics[width=\textwidth]{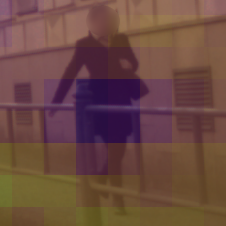}
\caption{{$\sigma_w^2$ aleatoric}}
\end{subfigure}
\begin{subfigure}{.13\textwidth}
\includegraphics[width=\textwidth]{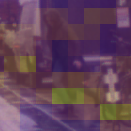}
\caption{{$\sigma_h^2$ epistemic}}
\end{subfigure}

\begin{subfigure}{.13\textwidth}
\includegraphics[width=\textwidth]{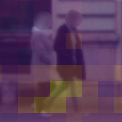}
\caption{{$\sigma_y^2$ epistemic}}
\end{subfigure}
\begin{subfigure}{.13\textwidth}
\includegraphics[width=\textwidth]{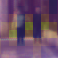}
\caption{{$\sigma_x^2$ epistemic}}
\end{subfigure}
\begin{subfigure}{.13\textwidth}
\includegraphics[width=\textwidth]{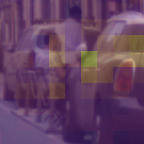}
\caption{{$\sigma_x^2$ epistemic}}
\end{subfigure}
\begin{subfigure}{.13\textwidth}
\includegraphics[width=\textwidth]{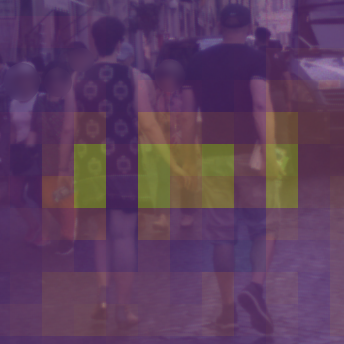}
\caption{{$\sigma_x^2$ aleatoric}}
\end{subfigure}
\begin{subfigure}{.13\textwidth}
\includegraphics[width=\textwidth]{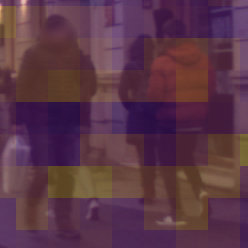}
\caption{{$\sigma_h^2$ epistemic}}
\end{subfigure}
\begin{subfigure}{.13\textwidth}
\includegraphics[width=\textwidth]{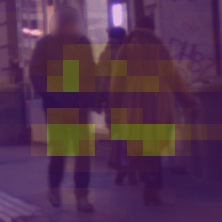}
\caption{{$\sigma_y^2$ aleatoric}}
\end{subfigure}
\begin{subfigure}{.13\textwidth}
\includegraphics[width=\textwidth]{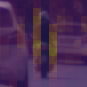}
\caption{{$\sigma_x^2$ aleatoric}}
\end{subfigure}

\caption{Uncertainty estimations visualized. Bright yellow indicates high uncertainty compared to the rest of the image. Note the low uncertainty at the center of each object for the regression variances.}
\label{fig:qa}

\end{figure*}

%% file: conclusion_and_acknowledgments.tex
\section{Conclusion}
\label{sec:conclusion}

In this work we showed how to incorporate the methods proposed by \cite{Gal2016Bayesian}, \cite{kendall_what}, and \cite{Gal2016thesis} into a modern one-stage object detection framework. We showed how to approximate the variational predictive distribution of our model and how to capture the different uncertainty measures. We provided the code to train models with epistemic and aleatoric uncertainties in TensorFlow. Along with a re-implementation of YOLOv3 in TensorFlow which supports training from scratch.

We showed in a quantitative and qualitative analysis that the uncertainty measure correlates with the overlap with objects. Predictions which are slightly off have higher uncertainty than more precise ones, which is exactly what we would expect from a sensible uncertainty estimation. This also means the uncertainty estimation for the bounding box regression could be used to improve non-maximum suppression, by favouring boxes with lower bounding box regression variance. We also found the aleatoric uncertainty to be correlated with the occlusion of objects making it a measure of problem inherent ambiguities.

In future work we plan to calibrate the uncertainty estimation for different prior sizes to make them comparable to each other. A viable solution seems to be to incorporate the techniques proposed by \cite{Phan2018calibrate} to calibrate the bounding box regression uncertainty. Also we want to address the effect of adverse weather situations such as rain or fog on the uncertainty measure. This work provides us with the necessary framework and theory to further research uncertainty estimations in object detection tasks.

\section*{ACKNOWLEDGMENT}
The research for this article has received funding from the European Union under the H2020 ECSEL Programme as part of the DENSE project, contract number 692449.